\documentclass[letterpaper,10pt,conference]{ieeeconf}
\IEEEoverridecommandlockouts       
\usepackage{amssymb}
\usepackage{amsmath}
\usepackage[english]{babel}
\usepackage{float}
\usepackage{graphicx}
\usepackage{hyperref}
\usepackage{mathtools}

\usepackage[noadjust]{cite}
\usepackage[font=footnotesize,labelfont=footnotesize]{caption}
\usepackage{comment}
\usepackage{authblk}
\usepackage{multirow}
\usepackage{xcolor}
\usepackage{algorithm}
\usepackage{algorithmic}
\usepackage{booktabs}
\usepackage{array}

\definecolor{rolecolor}{HTML}{D2242C}   
\definecolor{taskcolor}{HTML}{008600}   
\definecolor{systembar}{HTML}{F4C3C1}      
\definecolor{humanbar}{HTML}{BDDABA}      
\definecolor{promptbg}{HTML}{F7F7F7}      
\definecolor{prompttitlebg}{HTML}{F4C3C1} 

\newcommand{\ptcolor}[1]{\textcolor{rolecolor}{\textbf{#1}}}
\newcommand{\ptcolorq}[1]{\textcolor{taskcolor}{\textbf{#1}}}

\usepackage[most]{tcolorbox}
\tcbset{colback=promptbg, colframe=black, boxrule=0.9pt, arc=2pt,
        left=6pt, right=6pt, top=6pt, bottom=6pt, breakable,
        fonttitle=\bfseries, colbacktitle=prompttitlebg, coltitle=black}

\newtcolorbox{promptbox}[2]{enhanced,
  title={#1},
  colbacktitle=#2
}

\newenvironment{SystemPrompt}{\begin{promptbox}{System}{systembar}\small}{\end{promptbox}}
\newenvironment{HumanPrompt}{\begin{promptbox}{Human}{humanbar}\small}{\end{promptbox}}

\newcommand\lsb[1]{{\color{black}{#1}}}
\definecolor{darkblue}{rgb}{0.15,0.15,0.55}
\definecolor{lightgrey}{rgb}{0.75,0.75,0.75}

\providecommand{\codecomment}[1]{\textcolor{lightgrey}{\dotfill}\textcolor{darkblue}{//\,\textrm{#1}}}

\bstctlcite{RAL:BSTcontrol}

\begin{document}
\title{\LARGE \bf HEART: Coordination of Heterogeneous Expert Agents for Physically Grounded Robotic Task Planning}
\author{Junho Lee$^{\dagger}$, Seabin Lee$^{\dagger}$, Wonjong Lee, Nayoung Kim, Moonjeong Kang, and Changjoo Nam$^{*}$
\thanks{This work was supported by the National Research Foundation of Korea (NRF) grant funded by the Korea government (MSIT) (RS-2024-00461583) and  by the Korea Planning \& Evaluation Institute of Industrial Technology (KEIT) grant funded by the Korea government (MOTIE) (RS-2024-00444344).
All authors are with the Dept. of Electronic Engineering, Sogang University, Seoul, Korea. $^\dagger$Equal contribution (\emph{Co-first authors}). $^*$Corresponding author: {\tt\small cjnam@sogang.ac.kr}. Code and data: \url{https://anonymous.4open.science/r/HEART-anonymous-178D}}
}

\maketitle

\begin{abstract}
Large Language Models (LLMs) can reason over complex instructions but often fail to satisfy the physical and spatial constraints required for robotic task planning. Recent LLM-based planners directly translate text into action sequences, yet they lack structured reasoning about feasibility, reachability, and logical order, resulting in invalid or incomplete plans. We present a heterogeneous multi-LLM framework that decomposes instructions into atomic reasoning tasks and allocates them to role-specialized expert agents under a token budget for real-world computational and communicational constraints. By combining role-oriented reasoning from heterogeneous agents followed by constraint-driven plan synthesis, HEART validates capability, reachability, and constraint conditions before planning \lsb{and helps produce} physically executable plans while maintaining efficiency. Experiments across different household benchmarks show that HEART consistently \lsb{improves plan success compared to single-LLM and rule-based planners}, demonstrating that heterogeneous LLM collaboration enables robust and scalable robotic task planning under resource constraints.


\end{abstract}

\section{Introduction}
\label{sec:introduction}
Large Language Models (LLMs) have recently gained attention in robotic task planning~\cite{rana2023sayplan, liu2025delta, Palm-e, liu2025coherent} as they can deal with natural language and reason without explicit symbolic domain models. This flexibility contrasts with classical planners~\cite{FFplanner}, which rely on predefined operators and often fail under ambiguous instructions. However, robotic task planning in physical environments must satisfy two conditions: logical validity and physical executability. Logical validity ensures consistent action order and preconditions, while physical executability ensures robot capabilities, reachability, and feasibility. Single LLMs often struggle to handle these requirements jointly, generating plans that appear reasonable in language but fail when executed on real robots. For example, when asked to \lsb{``place any single apple on the dining table,''} a single LLM may create a plausible plan (e.g., $navigate$, $pick$, $navigate$, $place$) but often does not consider whether the robot is holding the apple before executing the $place$ action (i.e., logically invalid) or whether the robot can grasp the apple with its mechanical design (i.e., physically inexecutable).

To generate task plans that are both valid and executable, reasoning should jointly consider capability, environment, path, feasibility, and constraint. However, single-agent LLM approaches struggle to integrate these aspects because all reasoning must occur within a single monolithic context. This forces prompts to include extensive physical descriptions, causing context lengths to grow unnecessarily. Large contexts not only increase computational latency, memory usage, and inference cost, but also degrade reasoning quality due to diluted attention over long sequences~\cite{liu-etal-2024-lost}. A naïve multi-LLM approach, where agents reason in parallel or sequence without coordination, still fails to produce consistent reasoning~\cite{mandi2024roco}. Each agent receives the same full context, leading to redundant processing, excessive token usage, and conflicting reasoning results that compromise plan validity and executability~\cite{guo2024large, chen2024scalable, chen2023multi}. 

Therefore, LLM-based robotic task planning faces two key challenges: achieving valid and executable plans through heterogeneous reasoning, and maintaining efficiency in multi-agent collaboration by allocating reasoning tasks to the most relevant agents while keeping communicated tokens within a reasonable budget~\cite{wang2025mixture, singh2025improving, tran2501multi}.

\begin{figure}
    \centering
    \includegraphics[width=1.0\linewidth]{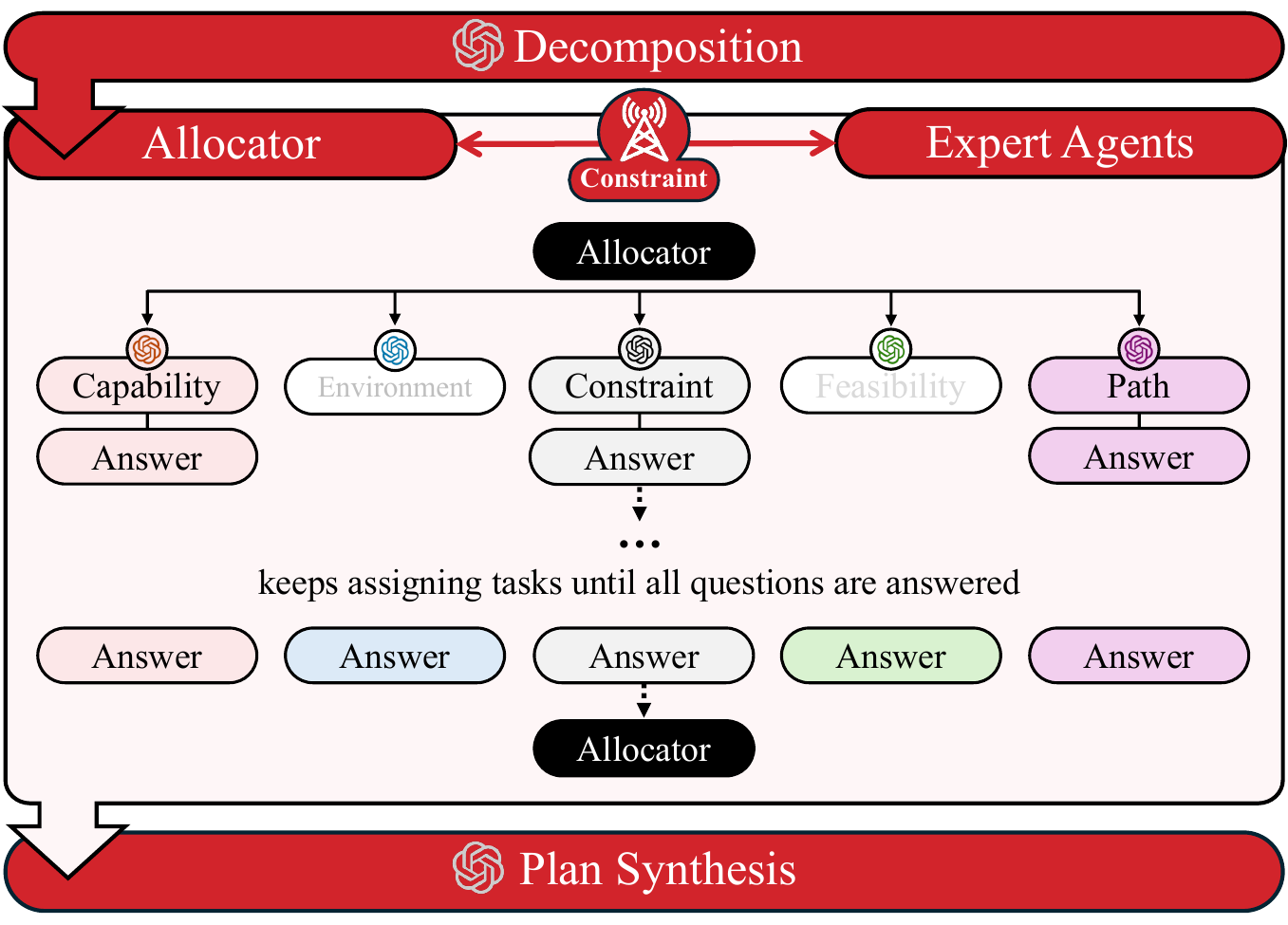}
    \caption{An overview of HEART. The framework decomposes the planning process into reasoning questions, assigns them to role-specialized expert agents through the allocator, and synthesizes their responses to generate a valid and physically executable plan.}
    \label{fig:overview}\vspace{-20pt}
\end{figure}

We propose \underline{H}eterogeneous \underline{E}xpert \underline{A}gents for physically grounded \underline{R}obotic \underline{T}ask planning (HEART), a coordinated multi-LLM framework that combines heterogeneous reasoning with token-budgeted task allocation for robotic planning. As shown in Fig.~\ref{fig:overview}, HEART decomposes a planning problem into atomic reasoning questions and assigns them to five role-specialized agents (i.e., \textit{capability}, \textit{environment}, \textit{path}, \textit{feasibility}, and \textit{constraint}) under a shared token budget. A centralized allocator coordinates this process by routing each question to the most relevant agent and distributing token budgets across roles to reduce redundancy and maintain efficiency, while constraining the use of memory and communication resources. Each agent focuses only on information relevant to its role, thereby improving reasoning precision, and their outputs are integrated in the planning stage \lsb{to help produce} valid and physically executable task plans. \lsb{HEART addresses these challenges by integrating heterogeneous reasoning with resource-efficient coordination for physically grounded and executable robotic task planning.}

\section{Related Work}
\label{sec:related} 
LLMs have been explored as planners that map natural language instructions to executable action sequences. Early approaches such as SayCan~\cite{saycan}, PaLM-E~\cite{Palm-e}, and SayPlan~\cite{rana2023sayplan} grounded instructions in affordances, multimodal perception, or scene graphs to generate plans linked to the real environment. Subsequent studies, including SMART-LLM~\cite{kannan2024smart}, extended this idea to multi-robot settings but still faced limited precision and reasoning depth~\cite{valmeekam2022large, pallagani2024prospects}. These limitations motivated LLM-as-translator frameworks that convert planning problems into formal representations such as PDDL~\cite{liu2025delta, fox2003pddl2} or temporal logics~\cite{wu2025selp, Autotamp}. These LLM-as-translator approaches improve logical consistency from symbolic solvers and validation tools. However, because the translation step separates symbolic reasoning from physical and spatial context, these methods still struggle to capture diverse reasoning aspects such as spatial coordination, physical feasibility, and temporal constraints, often resulting in failed execution~\cite{Autotamp,chen2024solving, li2024challenges}.

Recent work has also explored multi-agent LLM frameworks for robotic task planning, where multiple agents collaborate to extend reasoning consistency in planning. RoCo~\cite{mandi2024roco} assigns one agent per robot and coordinates task and motion planning through dialogue and feedback. Other frameworks~\cite{TAPAS, cao2024llm} focus on collaborative problem formulation and task decomposition, where multiple LLMs assume complementary roles to jointly construct domain models and refine planning steps. Multi-agent LLM systems have also been proposed for policy generation and control synthesis, such as MAS for robotic autonomy~\cite{chen2025multi} and Triple-S~\cite{triple-S} for long-horizon policy design. A related approach, FlowPlan~\cite{lin2025flowplan} uses LLMs with multi-step logical flows to capture operational constraints in task instructions. Despite these advances, most existing multi-agent LLM frameworks still lack explicit mechanisms for coordinating reasoning under communication constraints~\cite{wang2025mixture}. Without efficient task allocation, agents often duplicate reasoning over overlapping context, reach inconsistent results, and 
\lsb{may become inefficient} to scale when token budgets are limited. These limitations show the need for a systematic allocation strategy that distributes reasoning questions efficiently and maintains consistency across agents operating within communication limits~\cite{tran2501multi, chen2023multi, chen2024scalable}.

Beyond LLM-based methods, resource allocation has long been a central issue in multi-robot and distributed systems, from classical task assignment and market-based optimization~\cite{gerkey2004formal,dias2006market,korsah2013comprehensive} to adaptive scheduling in heterogeneous clusters~\cite{topcuoglu2002performance}. \lsb{In developmental robotics, socially guided intrinsic motivation frameworks address a related problem in which a robot selectively relies on external guidance under limited interaction budgets, and in some settings actively chooses among multiple teachers and learning strategies~\cite{nguyen2014socially,nguyen2012active}. This line of work provides a useful conceptual precedent for HEART, whose allocator routes reasoning subtasks to role-specialized agents under a token budget.} Recent compute-allocation studies in LLM inference~\cite{singh2025improving,chen2024scalable,snell2024scaling} further show that careful budgeting of reasoning resources \lsb{can improve efficiency and robustness, reinforcing the rationale for our token-aware allocation strategy.}

\begin{figure*}
    \centering
    \includegraphics[width=1\textwidth]{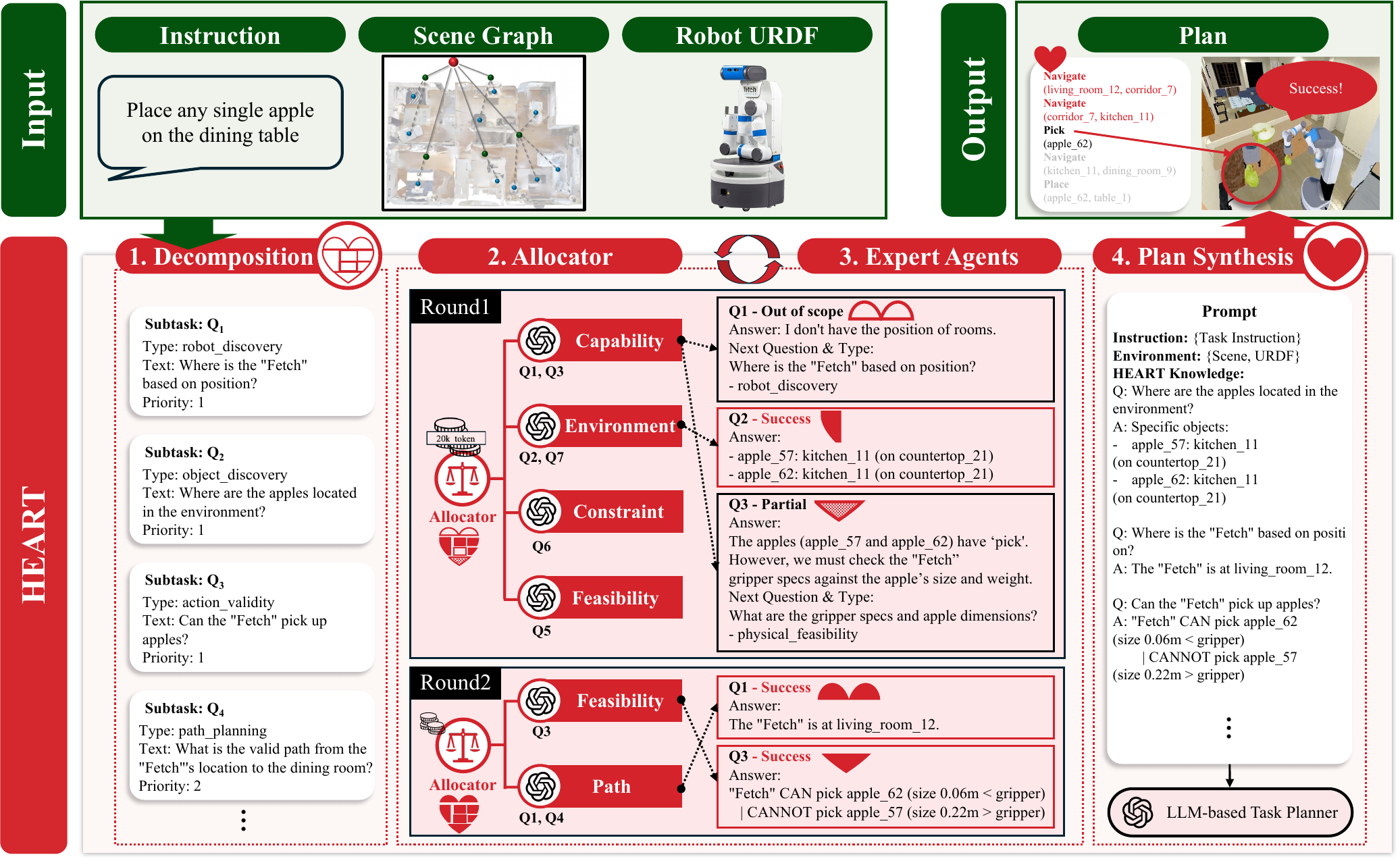}
    \caption{Overview of the HEART pipeline. Natural language instructions, scene graphs, and robot URDFs are decomposed into subtasks, allocated to heterogeneous expert agents under a token budget, and resolved through role-aligned reasoning. Agent responses (e.g., object discovery, feasibility checks) are integrated into planner inputs, producing executable task plans. The example shows a Fetch robot performing \lsb{\textit{``Place any single apple on the dining table''}}, decomposed into subtasks (Q1–Q7) and synthesized into a valid, executable plan.}
    \label{fig:framework}   
    \vspace{-10pt}
\end{figure*}

\section{Problem Statement}
\label{sec:prob}
We consider the problem of LLM-based, instruction-driven robot task planning, where natural language instructions should be mapped to task plans that are both valid and executable. The input consists of high-level instruction $I$ and environment data $E = (R, G)$ where $R$ includes robot specifications   and current states and $G$ is a 3D scene graph~\cite{armeni20193d} describing the environment with rooms, connectivity, and objects with affordances and physical properties. The output is a plan $\pi$, expressed as a sequence of symbolic task actions, that is \textit{valid} (i.e., satisfying logical and temporal constraints of the task) and \textit{executable} (i.e., respecting robot capabilities and physical feasibility).

The challenge is twofold: (i) plans must consider robot capability, physical and spatial feasibility, and logical order of actions to ensure the validity and executability; (ii) reasoning must consider limited computational and memory resources, since expanding the context size increases computational cost, memory usage, and inference latency. Thus, we introduce a per-round token budget $b$ that restricts the amount of communicated tokens per each reasoning. Our objective is to produce valid and executable task plans while maintaining efficiency under the token budget.

While this problem applies broadly across robotic application domains, we focus on household mobile manipulation tasks as a running example throughout the paper. In this example, robots must navigate and manipulate objects (e.g., pick up, open, turn on) in structured indoor environments, which are general capabilities across varied domains.

\section{Methods}
\label{sec:method}

We develop HEART, a token-aware multi-LLM framework for robotic task planning that coordinates heterogeneous expert agents through a central allocator. The HEART pipeline consists of four stages, illustrated in Fig.~\ref{fig:framework}.
(i) \textbf{Task decomposition} translates the instruction into atomic reasoning questions, which are derived from the environment data by identifying what additional information is required to accomplish the task. We regard these questions as subtasks, each resolved through its corresponding answer.
(ii) \textbf{Task allocation} assigns these subtasks to expert agents under a per-round token budget $b$.
(iii) \textbf{Heterogeneous expert reasoning} executes the assigned subtasks in parallel, with each agent operating on role-aligned and filtered data. Stages (ii)–-(iii) repeat until all subtasks are resolved or a termination condition is met.
(iv) \textbf{Plan synthesis} integrates the resulting Q\&A outcomes \lsb{into the planner input for generating} valid and executable task plans.

\begin{table}[t]
\footnotesize
\caption{\lsb{Summary of the 11 reasoning task types. The rightmost column shows the TypeBased baseline mapping (Sec.~\ref{sec:exp_allocation}).}}
\label{tab:types_schema_mapping}
\begin{tabular}{p{2.48cm} p{3.35cm} p{1.46cm}}
\toprule
\multirow{2}{*}{\textbf{Task Type}} & \multirow{2}{*}{\textbf{Brief Description}} & \textbf{Typebased} \\
& & \textbf{Agent}\\
\midrule
Robot Capability 
& Identify Robot Capability 
& Capability \\
Action Validity 
& Check Action-Object Match
& Capability \\
Object Discovery 
& Find Objects in Scene
& Environment \\
Scene Understanding 
& Analyze Spatial Relationship
& Environment \\
Robot Discovery 
& Determine Robot Location 
& Path \\
Path Planning 
& Find Valid Navigation Path 
& Path \\
Route Optimization 
& Select Most Efficient Path 
& Path \\
Physical Feasibility 
& Calculate Robot Physical 
& Feasibility \\
Task Dependency 
& Extract Action Prerequisites 
& Constraint \\
Action Constraint
& Derive Partial Order Plan 
& Constraint \\
Contextual Constraint
& Identify Implicit Rules
& Constraint \\
\bottomrule
\end{tabular}
\vspace{-15pt}
\end{table}

\subsection{Task Decomposition}
\label{sec:decompose} 

The first stage of HEART decomposes a high-level instruction $I$ into atomic reasoning subtasks $\mathcal{Q}=\{q_i\}_{i=1}^N$. \lsb{These subtasks cover five reasoning dimensions --- capability, environment, path, feasibility, and constraint.} Each subtask is labeled with one of the 11 task types defined in Table~\ref{tab:types_schema_mapping}, drawn from the commonly required reasoning needs across recent LLM-based robotic planning works~\cite{liu2025coherent,Palm-e,chen2025multi,mandi2024roco, saycan, kannan2024smart, wu2025selp}. \lsb{These types are domain-configurable labels that users can adapt without modifying the architecture.}

\lsb{The LLM-based decomposer receives the instruction $I$, the environment data $E$, and the TaskType definitions, and produces a list of subtasks where each is defined} 
as $q_i= (x_i,r_i,$ $\tau_i,p_i,stat_i)$. $x_i$ is the reasoning question, $r_i$ the answer, $\tau_i\!\in\!\mathcal{T}$ the task type, $p_i$ a local priority \lsb{(with $1$ the highest, reflecting execution dependency and task-level importance)}, and \lsb{$stat_i \in \{\texttt{SUCCESS}, \texttt{PARTIAL}, \texttt{OUT\_OF\_SCOPE}\}$ the status flag}. \lsb{The number of subtasks adapts to task complexity (typically $5$--$20$ in our evaluation), and the decomposer follows explicit guidelines (e.g., data-aware questions that avoid unavailable data); full prompts are released in the code repository.} During expert reasoning (Sec.~\ref{sec:expert-reasoning}), subtasks are updated: $r_i$ is populated, $x_i$ or $\tau_i$ can be updated, and $stat_i$ is set to \texttt{SUCCESS}, \texttt{PARTIAL}, or \texttt{OUT\_OF\_SCOPE}.

In our example (Fig.~\ref{fig:framework}), the instruction ``place any single apple on the dining table'' is decomposed into subtasks for the robot location (Q1, robot discovery), object positions such as apples and the dining table (Q2, object discovery), and pick-and-place capability (Q3, action validity).


\subsection{Allocation under Token Budget}
\label{sec:allocation}
The allocator centrally maps subtasks from $\mathcal{Q}$ to expert agents $\mathcal{A}$ under a per-round token budget $b$, which models limited computational and memory resources.\footnote{We assume that the value of $b$ is given as calculating an exact value for a specific system is out of scope.} Each LLM agent $a \in \mathcal{A}$ is represented by $a=(d_a, F(a,E), c_a)$, where $d_a$ is its role description, $F(a,E)$ is the role-aligned filtered environment data, and $c_a$ is an estimated token usage based on prompt size and past query history. This representation allows the allocator to match subtasks to suitable agents and approximate token consumption during scheduling.

This stage, described in Alg.~\ref{alg:allocator}, addresses two key requirements:  
(i) Subtask assignment, which combines semantic relevance with history-sensitive rescheduling of unresolved subtasks; and  
(ii) Capacity planning, which ensures the token budget while maintaining balanced utilization across agents.  

\subsubsection{Subtask assignments with history-sensitive penalty}
\label{sec:task-agent-assign}
For each subtask $q_i$ with reasoning question $x_i$ and task type $\tau_i$, the allocator computes a semantic score over agents using text embeddings from Sentence-BERT~\cite{sentence-transformer}:
\[
s_i(a) \;=\; \cos\!\big(e(x_i), e(d_a)\big) \;+\; \cos\!\big(e(\tau_i), e(d_a)\big) \;-\; \Delta_i(a),
\]
where $d_a$ is the role description of the agent $a$, $e(\cdot)$ denotes the embedding function, and $\Delta_i(a)$ is a history-sensitive penalty that increases with previous failed attempts on $a$ and with the most recent failed agent. Without this penalty, a subtask may be repeatedly assigned to agents that have already failed. The selected agent is $\varphi(q_i) = \arg\max_{a \in \mathcal{A}} s_i(a)$. To avoid unbounded retries, each subtask is assigned a maximum trial limit (set to five in our experiments). Subtasks that exceed this limit without achieving $stat_i=\texttt{SUCCESS}$ are marked as \texttt{FAIL} and excluded from further allocations.

\begin{algorithm}[t]
\caption{\textsc{Allocator}}
\label{alg:allocator}
\begin{algorithmic}[1]
\STATE \textbf{Input:} Subtasks $\mathcal{Q}$, expert agents $\mathcal{A}$, budget $b$
\STATE \textbf{Output:} Assignment set $M = \{(q_i, \varphi(q_i))\}$
\FOR{each $q_i \in \mathcal{Q}$} \label{line:start_for1}
  \STATE Compute $s_i(a)$ with embeddings and history penalty 
  \STATE $\varphi(q_i) \gets \arg\max_{a \in \mathcal{A}} s_i(a)$ \codecomment{Sec.~\ref{sec:task-agent-assign}} 
\ENDFOR \label{line:end_for1}
\STATE Compute demand for each $a$ from $\{\varphi(q_i) \mid q_i \in \mathcal{Q}\}$\label{line:demand}
\STATE Plan capacities $y_a$ s.t. $\sum_{a \in \mathcal{A}} y_a \, c_a \le b$ \label{line:capacity} \codecomment{Sec.~\ref{sec:capa-plan}}
\STATE Select up to $y_a$ subtasks per agent by priority $p_i$ \label{line:filter}
\STATE $M \gets \{(q_i, \varphi(q_i)) \mid q_i \text{ dispatched}\}$; update history\label{line:collect} 
\RETURN $M$
\end{algorithmic}
\end{algorithm}

\subsubsection{Capacity planning with token budget}
\label{sec:capa-plan}
From the subtask assignment, we compute the demand for each agent type based on the remaining unresolved subtasks. Using estimated token count $c_a$, the allocator computes capacities $y_a$ under the budget $b$ (in tokens):
\[
\sum_{a \in \mathcal{A}} y_a\,c_a \;\le b,\qquad
y_a \in \mathbb{Z}_{\ge 0}.
\]

We adopt a greedy allocation strategy that iterates over agents in decreasing order of token count $c_a$, assigning as many instances as possible until the demand or budget is exhausted. If $c_a > b$, only one instance of $a$ is allowed. Subtasks are then scheduled by priority $p_i$, and the allocator outputs an assignment set $M = \{(q_i, \varphi(q_i))\}$ for the round. The estimated token count is updated after each round to refine future capacity planning. Failed attempts are also recorded for each subtask (line~\ref{line:collect}).

In our example (Fig.~\ref{fig:framework}), subtasks such as Q1, Q2, and Q3 are initially routed by semantic similarity, Q4 is deferred under the budget, and Q1 and Q3 are reassigned after returning partial or out-of-scope answers.


\subsection{Heterogeneous Expert Reasoning}
\label{sec:expert-reasoning}
The third phase executes allocated subtasks using heterogeneous expert agents. Each agent specializes in a distinct reasoning aspect required for robotic task planning. To support role specialization, each agent receives only the role-aligned filtered data $F(a,E)$ introduced earlier, rather than the full scene. \lsb{This filtering, applied during agent reasoning rather than at the planner stage,} reduces context size and redundancy, improves efficiency and reasoning quality, and aligns with reports that modular, specialized models are more economical and robust than monolithic ones~\cite{belcak2025small}.

Given an assigned subtask $q_i=(x_i,r_i,\tau_i,p_i,stat_i)$, agent $a$ processes it using the reasoning question $x_i$ together with $F(a,E)$ and returns an updated answer $r_i$ with status flag $stat_i\in\{\texttt{SUCCESS},\texttt{PARTIAL},\texttt{OUT\_OF\_SCOPE}\}$. Agents operate in parallel within each round as determined by the allocator; some may process multiple subtasks while others remain idle. The five instantiated roles are:

\textit{Capability Reasoner} evaluates what the robot can do in its current configuration. Using URDF and robot state (kinematics, joint limits, base mobility, gripper type, predefined action set), it determines whether specified navigation or manipulation actions are supported for given objects and what interaction modes are allowed.

\textit{Environment Reasoner} focuses on scene and object understanding. From the information provided by 3D scene graph (e.g., rooms, objects, affordances, accessibility flags, container relations), it identifies relevant objects, their states, and spatial context needed by subsequent reasoning.

\textit{Path Reasoner} plans high level navigation sequences based on the connectivity between spaces. Using room adjacency, spatial coordinates, and the current robot pose, it decides whether the required locations are mutually reachable and returns room-sequence routes that respect adjacency.

\textit{Feasibility Reasoner} validates the physical feasibility of candidate actions. It combines the kinematic information of the robot (e.g., arm length, gripper size) of the robot with object properties (e.g., dimensions, weight, material) to judge graspability, reachability, and payload limits for the action.

\textit{Constraint Reasoner} extracts logical and temporal constraints from the instruction and environment. Using affordances, object states, and accessibility information, it derives ordering relations (e.g., open the gripper, pick an object, and close the gripper, resulting in an action sequence $open, pick, close$) and resolves ambiguities to produce action dependencies for a consistent plan.

All expert agents follow the same prompting schema to standardize inputs and outputs across roles. The general template is shown below, where role-level fields (fixed per agent) are highlighted in red, and subtask-specific fields (varying per subtask) are highlighted in green\lsb{; full system and human prompts (with agent-specific decision rules) are released in the code repository}.

\begin{SystemPrompt}
You are a \ptcolor{Specific-Agent} specialized in robotic task planning.
Your primary role is to \ptcolor{Specific-Agent Purpose}, supported by a role-aligned view of the environment that provides \ptcolor{Specific-Agent-Aligned Data Description}.
In this role, you reason about \ptcolor{Specific-Agent Reasoning}, following \ptcolor{Specific-Agent Rules} that constrain your scope.
Through this analysis you contribute \ptcolor{Specific-Agent Contribution}, which is integrated into the broader planning process.
\end{SystemPrompt}

\begin{HumanPrompt}
\textbf{\#\# Task}\\
Question: \ptcolorq{Question}\\
Current Focus: \ptcolorq{Question Type}\\
Original Goal: \ptcolorq{Original Question}\\
Original Type: \ptcolorq{Original Question Type}

\medskip
\textbf{\#\# Agent Data}\\
Your Data: \ptcolor{Specific-Agent-Aligned Data}\\
Previous Context: \ptcolor{Specific-Agent Memory Data}

\medskip
\textbf{\#\# Decision Guidelines}\\
Can you answer the ORIGINAL question in the required format? $\rightarrow$ \texttt{SUCCESS}\\
Can you contribute ANY useful information? $\rightarrow$ \texttt{PARTIAL}\\
Is it completely outside your expertise with no relevant data? $\rightarrow$ \texttt{OUT\_OF\_SCOPE}

\medskip
\textbf{\#\# Response Format}\\
Status: \texttt{SUCCESS} / \texttt{PARTIAL} / \texttt{OUT\_OF\_SCOPE}\\
Answer: Based only on verified data; include any details helpful to the next agent.\\
Follow-up Question: (if not \texttt{SUCCESS}) Minimal extra info needed, phrased for the appropriate agent.\\
Follow-up Task Type: (if not \texttt{SUCCESS}) Choose from available task types.
\end{HumanPrompt}

Each response of the agents updates $r_i$ and sets $stat_i$ according to the completion criteria of the current type $\tau_i$: \texttt{SUCCESS} when all required fields are present, \texttt{PARTIAL} when only a subset is available, and \texttt{OUT\_OF\_SCOPE} when no role-relevant inference is possible. For \texttt{PARTIAL} or \texttt{OUT\_OF\_SCOPE}, the agent rephrases $x_i$ and retags $\tau_i$ to propose a minimal follow-up for later rounds.

Completion of response is type-specific: the prompt enumerates the required fields for each $\tau_i$ (e.g., \textit{capability} requires an \texttt{action\_set}, \textit{path} requires \texttt{valid\_paths}, and \textit{feasibility} requires boolean checks of specified physical properties). Each agent maintains a local memory of its prompts and responses, \lsb{which is re-injected in later rounds to maintain consistency across retries}.

In our example (Fig.~\ref{fig:framework}), Q1 returns \texttt{OUT\_OF\_SCOPE} from the Capability Reasoner (location lies outside its role) and Q3 returns \texttt{PARTIAL} (graspability needs object information from other agents); both are reassigned, while Q2 is directly resolved by the Environment Reasoner as \texttt{SUCCESS}.


\subsection{Plan Synthesis}
\label{sec:synthesis}
The final phase of HEART \lsb{produces the planning context that a task planner consumes to create a valid and executable plan} $\pi$. Each resolved subtask $q_i=(x_i,r_i,\tau_i,p_i,stat_i)$ contributes facts or constraints, and the collection of $(x_i,r_i,\tau_i)$ forms this context. \lsb{Synthesis cross-validates these Q\&A pairs, preserving partial answers from failed subtasks so that partial information is retained, and resolves contradictions between agents by treating a failed physical check (e.g., size or reach) as overriding a positive capability claim from another agent.} Following the general Q\&A reasoning paradigm explored in LLM workflows~\cite{yue2025survey}, HEART integrates the \lsb{resulting Q\&A block} into the planner prompt or problem specification, so that \lsb{the planner takes} capability limits, feasibility results, discovered objects, and ordering relations \lsb{into account during plan generation}. HEART can work with any LLM-based or hybrid planner that accepts natural language or structured textual input; if no plan is possible, the system returns an explicit infeasibility report.

In our example (Fig.~\ref{fig:framework}), synthesis cross-validates the resolved Q1 (location), Q2 (apples and their states), and Q3 (graspable apples) into a planning context, which the planner then uses to select a feasible target apple and generate a valid and executable plan.


\section{Experiments}
\label{sec:exp}

We design three experiments to evaluate HEART across different aspects of reasoning and performance:
(i) \textbf{\lsb{agent role ablation}}, testing whether role specialization improves planning robustness and efficiency \lsb{by varying the number of expert roles};
(ii) \textbf{\lsb{allocator ablation}}, comparing HEART \lsb{against baselines and ablating its components} under varying budgets; and
(iii) \textbf{task planning performance}, evaluating whether HEART improves the validity and executability of plans across both LLM-based and symbolic planners.

\subsection{Experimental Setup}
\subsubsection{Dataset}
\begin{figure}
    \centering
    \includegraphics[width=0.9\linewidth]{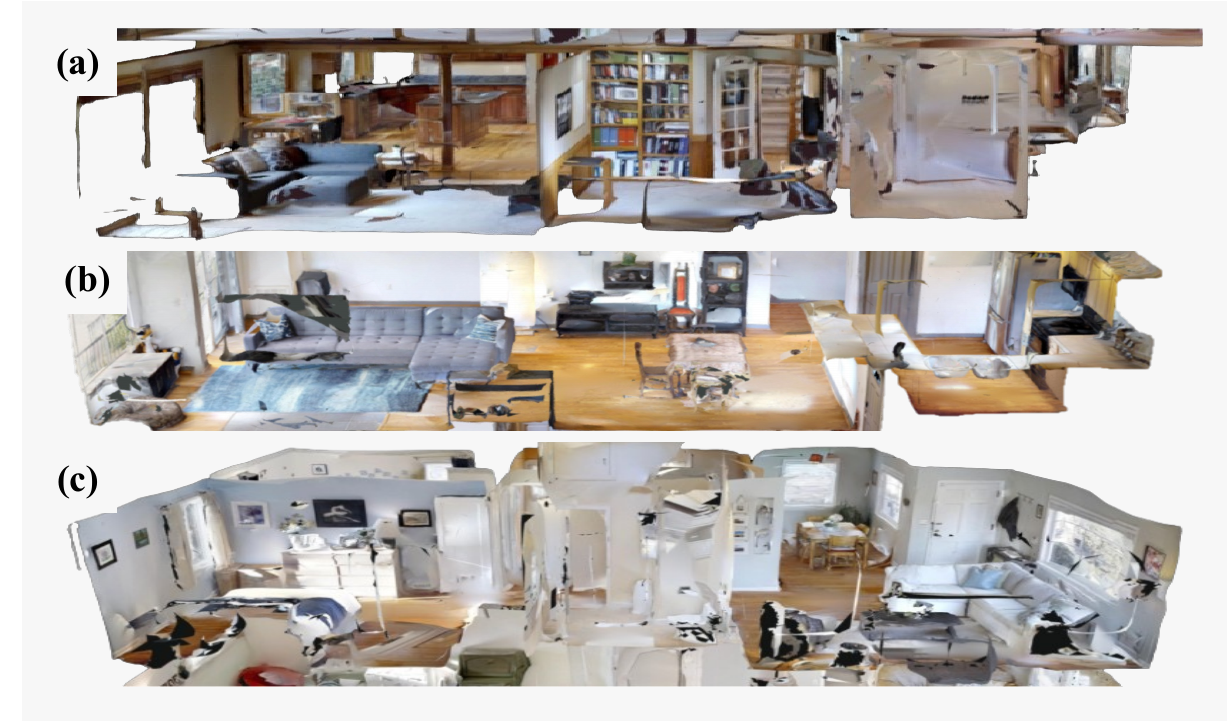}
    \caption{\lsb{Household manipulation scenes used for evaluation:
(a) \textbf{Beechwood} (9 rooms, 74 objects);
(b) \textbf{Benevolence} (5 rooms, 55 objects); 
(c) \textbf{Merom} (7 rooms, 65 objects).
Adapted from the 3D Scene Graph dataset~\cite{armeni20193d}.}}
    \label{fig:simulation}
\vspace{-15pt}
\end{figure}

We evaluate HEART on three household manipulation scenes from Gibson dataset~\cite{li2021igibson}, each chosen to emphasize different reasoning challenges. \textbf{Beechwood} (Fig.~\ref{fig:simulation}a) contains 9 rooms and 74 objects (15 affordances) and emphasizes logical sequencing. For example, tasks such as laundry (requiring opening the washer before loading clothes) and kitchen safety (e.g., turning off an active stove) involve clear preconditions and ordered action dependencies. \textbf{Benevolence} (Fig.~\ref{fig:simulation}b) consists of 5 rooms and 55 objects (11 affordances). It stresses physical feasibility: oversized apples that cannot be grasped, heavy sunglasses exceeding robot payload, and objects located at unreachable heights require URDF-based reasoning over robot kinematic constraints and payloads. \textbf{Merom} (Fig.~\ref{fig:simulation}c) includes 7 rooms and 65 objects (16 affordances). It is dedicated to multi-robot cooperation: all 10 tasks involve two robots, with both heterogeneous teams (\lsb{a mobile manipulator and a quadrotor}) and homogeneous teams (two mobile manipulators). Tasks combine ground-level manipulation and aerial inspection, requiring allocation across robots with complementary capabilities. \lsb{In total, we evaluate on 40 task instructions spanning these challenges (Beechwood 15, Benevolence 15, Merom 10). Ground-truth optimal plan lengths span $2$ to $18$ steps, with per-scene medians of $8$, $6$, and $10$ respectively, providing a difficulty proxy across short and long tasks. The full instruction list and ground-truth PDDL files are released in the code repository.}
\subsubsection{Evaluation Metrics}

We evaluate HEART using metrics that capture both efficiency and planning quality:
\begin{itemize}
    \item \textit{Rounds}: Number of allocation cycles under budget $b$, \lsb{reflecting the time spent in the allocator and agent stages}
    \item \textit{\lsb{Token count (k)}}: \lsb{Reported separately for allocation and reasoning (\textit{Tokens}), for the task planner only (\textit{Plan Tok.}), and for the end-to-end pipeline (\textit{Total Tok.})}
    \item \textit{SubSR}: Subtask success rate
    \item \textit{Plan SR}: Success rate of task instructions \lsb{whose generated plan is accepted by VAL~\cite{val} against the ground-truth PDDL files released per task --- manually authored from the robot URDF and scene-graph measurements, with physically infeasible objects (oversized, out of reach) excluded from each problem file}
    \item \textit{\lsb{Step Ratio}}: \lsb{Generated plan length divided by the ground-truth optimal plan length}
\end{itemize}

\subsubsection{Default Settings}
Unless otherwise noted, the experiments are done with the following default settings.
HEART is implemented using LangGraph, with task allocation relying on Sentence-BERT embeddings and tiktoken for token estimation. We use an LLM-as-Planner with chain-of-thought prompting as the default planner across all experiments. We also evaluate HEART with an additional planner, DELTA, a hybrid LLM–PDDL planning system with symbolic validation, to test compatibility with both natural language and symbolic planning pipelines. \lsb{The evaluation spans three robot types: Fetch and JR2-Kinova mobile manipulators and a quadrotor, across single- and multi-robot tasks.}

\lsb{Each task instruction is repeated for $10$ seeds, with decomposition rerun independently per seed and per-condition trial counts specified in each table caption.} Experiments use a per-round token budget of $20$k. Subtasks have a retry limit of five; exceeding it sets \texttt{FAIL}. Results are reported as averages over seeds with $95$\% confidence intervals.

\subsubsection{LLM Configuration}
The decomposer agent uses reasoning model \texttt{o4-mini}, while all expert agents (capability, environment, path, feasibility, and constraint) \lsb{and the synthesis agent} use \texttt{gpt-4o}. \lsb{LLM-CoT (with chain-of-thought) and DELTA use \texttt{gpt-4o}.} All models use temperature 0.0.

\subsubsection{Hardware}
Experiments ran on Ubuntu 22.04 with AMD 7900X, 32GB RAM, RTX 4080 SUPER; LLM calls use the OpenAI API.

\begin{table}[t!]
\footnotesize
\centering
\setlength{\tabcolsep}{3.9pt}
\caption{\lsb{Agent role ablation under a fixed 20k token budget. 50 trials per cell; mean $\pm$ 95\% CI for Rounds and Tokens. Scenes: BW=Beechwood, BN=Benevolence, MR=Merom.}}
\label{tab:agent_role}
\begin{tabular}{lccccc}
\toprule
Scene & Agents & Rounds & Tokens (k) & SubSR (\%) & PlanSR (\%) \\
\midrule
\multirow{3}{*}{BW} & 5 & $\mathbf{5.1 \pm 0.4}$  & $110.8 \pm 32.0$ & $97.4$           & $\mathbf{72.0}$ \\
                    & 3 & $5.2 \pm 0.5$           & $109.6 \pm 34.3$ & $98.7$           & $56.0$ \\
                    & 1 & $11.7 \pm 0.8$          & $158.7 \pm 37.4$ & $\mathbf{100.0}$ & $60.0$ \\
\midrule
\multirow{3}{*}{BN} & 5 & $\mathbf{5.1 \pm 0.4}$  & $111.1 \pm 32.1$ & $99.2$           & $\mathbf{76.0}$ \\
                    & 3 & $\mathbf{5.1 \pm 0.4}$  & $105.4 \pm 21.1$ & $98.6$           & $44.0$ \\
                    & 1 & $8.7 \pm 0.5$           & $147.9 \pm 23.5$ & $\mathbf{100.0}$ & $56.0$ \\
\midrule
\multirow{3}{*}{MR} & 5 & $6.9 \pm 0.6$           & $143.8 \pm 43.9$ & $97.1$           & $\mathbf{60.0}$ \\
                    & 3 & $\mathbf{6.5 \pm 0.6}$  & $140.4 \pm 42.5$ & $\mathbf{98.6}$  & $50.0$ \\
                    & 1 & $14.8 \pm 1.4$          & $197.5 \pm 63.9$ & $96.6$           & $56.0$ \\
\bottomrule
\end{tabular}
\vspace{-15pt}
\end{table}

\subsection{Agent Role Ablation}
\label{sec:exp_hete_homo}
\lsb{We compare three configurations: a 5-agent setting with five specialized agents (capability, environment, path, feasibility, constraint) and role-aligned filtered scene and robot data; a 3-agent setting with \emph{physical}, \emph{spatial}, and \emph{constraint} agents; and a 1-agent setting in which a single unspecialized agent receives the full unfiltered scene and robot data for every subtask. All three settings are designed independently rather than subsets of the 5-agent configuration, and use the same prompt format and per-round token budget.}

\lsb{Table~\ref{tab:agent_role} reports the results across the three scenes. The 5-agent configuration achieves the highest Plan SR on all three scenes ($72.0$\%, $76.0$\%, $60.0$\%). The 1-agent configuration reaches the highest Subtask SR ($100.0$\% on Beechwood and Benevolence) but its Plan SR is 12 to 20 percentage points below the 5-agent setting. The 3-agent configuration falls between the other two on Subtask SR but produces the lowest Plan SR ($56.0$\%, $44.0$\%, $50.0$\%). The 5-agent setting also uses about $25$ to $30$\% fewer total tokens than the 1-agent setting and finishes in roughly half the number of allocation rounds (\lsb{and therefore roughly half the time in the allocator and agent stages}), because a single agent receiving the full scene and robot data must use a large fraction of the per-round token budget for each subtask.}

\lsb{The pattern that 1-agent reaches the highest Subtask SR while 5-agent reaches the highest Plan SR shows that having an answer for every question is not the same as having an answer that is correct and sufficient for planning. A single agent with the full scene and robot data tends to answer feasibility questions from affordance information and skip physical checks against robot hardware, producing answers that lead to invalid plans. For example, in a pick-an-apple task with two apples ($0.228$\,m and $0.066$\,m, carrying the ``pick'' affordance) and a $0.1$\,m gripper, the single agent accepts the oversized apple based on affordance, while the 5-agent feasibility agent rejects it on diameter. 3-agent produces the lowest Plan SR because each agent spans more than one reasoning dimension, leaving inconsistencies that synthesis cannot resolve.}

\begin{table}[t!]
\footnotesize
\centering
\setlength{\tabcolsep}{5pt}
\caption{\lsb{Allocator ablation under different token budgets (Benevolence, 50 trials per cell; mean $\pm$ 95\% CI for SubSR, Rounds, and Tokens).}}
\label{tab:allocator_eff}
\begin{tabular}{llccc}
\toprule
Budget & Allocator & SubSR (\%) & Rounds & Tokens (k) \\
\midrule
\multirow{6}{*}{10k} & TypeBased             & $87.9 \pm 3.0$          & $12.0 \pm 0.8$          & $109.0 \pm 6.8$ \\
                     & LLM                   & $79.6 \pm 3.3$          & $10.4 \pm 0.7$          & $125.9 \pm 6.6$ \\
                     & \textbf{HEART (full)} & $99.0 \pm 0.7$          & $\mathbf{8.9 \pm 0.6}$  & $107.4 \pm 5.9$ \\
                     & w/o penalty           & $86.8 \pm 4.0$          & $9.6 \pm 0.8$           & $\mathbf{105.8 \pm 7.8}$ \\
                     & w/o capacity          & $\mathbf{99.2 \pm 0.8}$ & $10.4 \pm 0.7$          & $107.6 \pm 5.8$ \\
                     & w/o both              & $88.3 \pm 3.0$          & $11.5 \pm 0.7$          & $106.8 \pm 6.2$ \\
\midrule
\multirow{6}{*}{20k} & TypeBased             & $88.7 \pm 3.0$          & $6.3 \pm 0.5$           & $124.0 \pm 8.7$ \\
                     & LLM                   & $77.0 \pm 3.6$          & $6.5 \pm 0.5$           & $131.0 \pm 8.9$ \\
                     & \textbf{HEART (full)} & $\mathbf{99.5 \pm 0.5}$ & $\mathbf{5.0 \pm 0.4}$  & $\mathbf{108.8 \pm 7.0}$ \\
                     & w/o penalty           & $91.2 \pm 2.4$          & $6.0 \pm 0.6$           & $118.7 \pm 9.6$ \\
                     & w/o capacity          & $98.8 \pm 0.8$          & $5.1 \pm 0.4$           & $109.8 \pm 7.5$ \\
                     & w/o both              & $91.2 \pm 2.3$          & $6.3 \pm 0.5$           & $118.5 \pm 8.7$ \\
\midrule
\multirow{6}{*}{40k} & TypeBased             & $87.7 \pm 2.9$          & $4.9 \pm 0.3$           & $126.6 \pm 11.2$ \\
                     & LLM                   & $75.9 \pm 4.0$          & $5.1 \pm 0.3$           & $134.9 \pm 8.4$ \\
                     & \textbf{HEART (full)} & $\mathbf{98.9 \pm 0.9}$ & $\mathbf{3.5 \pm 0.3}$  & $108.2 \pm 7.4$ \\
                     & w/o penalty           & $90.3 \pm 2.3$          & $4.6 \pm 0.3$           & $118.3 \pm 8.6$ \\
                     & w/o capacity          & $98.8 \pm 0.9$          & $3.6 \pm 0.3$           & $\mathbf{107.1 \pm 7.4}$ \\
                     & w/o both              & $89.6 \pm 2.8$          & $4.6 \pm 0.3$           & $118.7 \pm 9.2$ \\
\bottomrule
\end{tabular}
\vspace{-15pt}
\end{table}

\subsection{Allocator Ablation}
\label{sec:exp_allocation}

\lsb{We compare HEART against two baseline allocators and three component ablations of HEART itself, across three token budgets ($10$k/$20$k/$40$k; $50$ trials per cell on Benevolence). The baselines are \textit{TypeBased} (rule-based mapping from each subtask type to a fixed agent) and \textit{LLM} (a general-purpose LLM that decides assignments). HEART combines semantic matching, a history-sensitive penalty, and capacity planning; the three ablations remove the penalty, capacity planning (which then falls back to random-order dispatch within the per-round budget), or both.}

\lsb{Table~\ref{tab:allocator_eff} reports the results. Across all three budgets, HEART reaches Subtask SR around $99$\% and uses the fewest allocation rounds at every budget. Total tokens are similar to the component ablations. The two components contribute differently: removing the history-sensitive penalty drops Subtask SR from $99.5$\% to $91.2$\% at 20k (rounds $5.0 \to 6.0$), since the penalty avoids re-dispatching a question to an agent that has already failed it. Removing capacity planning has a smaller effect at 20k but saves about $1.5$ rounds at the tightest 10k budget, since capacity planning schedules tasks per agent under the per-round budget so that higher-priority subtasks are not crowded out by random order.}

\lsb{Compared with the baselines, HEART maintains substantially higher Subtask SR with fewer tokens than LLM. The LLM baseline issues a routing call per new or re-routed question, raising total tokens by about $15$--$25$\% over HEART (Table~\ref{tab:allocator_eff}) without higher Subtask SR (around $77$\% at 20k). The rule-based mapping in TypeBased avoids that overhead but cannot adapt when subtask content does not match its type label, leaving Subtask SR around $88$\%. Across budgets, HEART therefore achieves high subtask success while avoiding the routing overhead of LLM allocation and the rigidity of fixed-type rules.}

\begin{table*}[t!]
\footnotesize
\centering
\setlength{\tabcolsep}{4pt}
\caption{\lsb{Task planning with LLM-CoT and DELTA planners, with and without HEART. BW/BN: 150 trials per cell; MR: 80 trials per cell. PlanSR CI computed as binomial proportion 95\% CI. Scenes: BW=Beechwood, BN=Benevolence, MR=Merom.}}
\label{tab:planner_eff}
\begin{tabular}{llcccccccc}
\toprule
\multirow{2}{*}{Scene} & \multirow{2}{*}{Config} & \multicolumn{4}{c}{LLM-CoT planner} & \multicolumn{4}{c}{DELTA planner} \\
\cmidrule(lr){3-6} \cmidrule(lr){7-10}
 & & PlanSR (\%) & Step Ratio & Plan Tok.\ (k) & Total Tok.\ (k) & PlanSR (\%) & Step Ratio & Plan Tok.\ (k) & Total Tok.\ (k) \\
\midrule
\multirow{2}{*}{BW} & w/o HEART        & $56.7 \pm 7.9$  & $1.16 \pm 0.16$         & $11.4 \pm 0.3$ & $11.4 \pm 0.3$   & $64.0 \pm 7.7$           & $1.07 \pm 0.09$         & $39.3 \pm 1.4$ & $39.3 \pm 1.4$ \\
                    & \textbf{+ HEART} & $77.3 \pm 6.7$  & $1.09 \pm 0.12$         & $12.5 \pm 0.5$ & $125.9 \pm 32.8$ & $\mathbf{86.0 \pm 5.6}$  & $\mathbf{1.07 \pm 0.09}$ & $45.6 \pm 2.1$ & $161.4 \pm 36.9$ \\
\midrule
\multirow{2}{*}{BN} & w/o HEART        & $16.7 \pm 6.0$  & $\mathbf{1.01 \pm 0.05}$ & $10.1 \pm 0.1$ & $10.1 \pm 0.1$   & $6.7 \pm 4.0$            & $1.17 \pm 0.17$         & $38.5 \pm 2.3$ & $38.5 \pm 2.3$ \\
                    & \textbf{+ HEART} & $60.7 \pm 7.8$  & $1.05 \pm 0.09$         & $11.2 \pm 0.3$ & $117.6 \pm 28.6$ & $\mathbf{74.0 \pm 7.0}$  & $1.09 \pm 0.18$         & $43.1 \pm 2.4$ & $148.9 \pm 28.6$ \\
\midrule
\multirow{2}{*}{MR} & w/o HEART        & $22.5 \pm 9.2$  & $1.44 \pm 0.15$         & $12.2 \pm 0.1$ & $12.2 \pm 0.1$   & $18.8 \pm 8.6$           & $1.32 \pm 0.38$         & $42.7 \pm 3.8$ & $42.7 \pm 3.8$ \\
                    & \textbf{+ HEART} & $65.0 \pm 10.5$ & $1.33 \pm 0.12$         & $13.3 \pm 0.3$ & $150.8 \pm 36.8$ & $\mathbf{76.2 \pm 9.3}$  & $\mathbf{1.07 \pm 0.11}$ & $46.8 \pm 2.7$ & $181.0 \pm 38.7$ \\
\bottomrule
\end{tabular}
\vspace{-15pt}
\end{table*}

\subsection{Task Planning with Different Planners}
\label{sec:exp_planner}

\lsb{We compare HEART + LLM-CoT and HEART + DELTA against each planner alone, on the same instructions and per-round budget (Table~\ref{tab:planner_eff}). The planners are LLM-CoT (an LLM that emits action sequences through chain-of-thought reasoning) and DELTA (an LLM-to-PDDL translation framework that solves PDDL with a symbolic planner). Without HEART, both planners reach limited Plan SR ($6.7$--$64.0$\% across scenes); HEART + LLM-CoT raises Plan SR to $60.7$--$77.3$\%, and HEART + DELTA reaches the highest values ($74.0$--$86.0$\%) with Step Ratio close to optimal.}

\lsb{The failure modes from VAL break down into three categories: precondition violation, goal not achieved, and parse/structural failure. The dominant baseline failure is a precondition violation from a hallucinated affordance or a wrong object selection. With HEART, the failure-relevant checks happen in the reasoning stage before plan generation: the feasibility agent rules out infeasible objects and the constraint agent rules out actions that violate ordering, so the planner generates the plan from already-validated context. This yields the larger Plan SR gains on Benevolence and Merom, where physical feasibility reasoning is most needed.}

\lsb{HEART increases total token consumption ($10$--$43$k $\to$ $117$--$181$k across both planners), the upfront cost of validating capability, feasibility, and constraint conditions before plan generation. This investment converts attempts into deployable plans, raising Plan SR from $6.7$--$64$\% (without HEART) to $60$--$86$\% across the three scenes. The planner-side cost (Plan Tok.\ in Table~\ref{tab:planner_eff}) rises only modestly ($\sim$$1$k for LLM-CoT, $\sim$$4$--$6$k for DELTA), and the reasoning-stage overhead can be further reduced by context pruning. The reasoning is agnostic to the planner: it enables both LLM-CoT and DELTA to generate valid and executable plans across all three scenes.}

\section{Conclusion}
\label{sec:conclusion}
We introduced HEART, a heterogeneous multi-agent system for robotic task planning. HEART decomposes natural language instructions into reasoning subtasks, allocates them to role-aligned expert agents under a per-round token budget, and integrates agent responses into plan synthesis. Across three household environments, HEART improves efficiency and plan success over homogeneous agents, baseline allocators, and single-planner baselines. \lsb{The validated reasoning context produced by HEART is used by both LLM-as-planner (LLM-CoT) and LLM-as-translator to a symbolic solver (DELTA), enabling each to generate valid and executable plans.} These results highlight the value of role specialization and allocation under per-round token budget for producing valid and executable task plans. \lsb{HEART produces valid and executable task plans that are robust and scalable, though at the cost of additional token usage.} This overhead can be mitigated through pruning and data reduction techniques, as mentioned in the experiments. Future work will extend the heterogeneity of HEART from the prompt level to the model level, where each expert role is paired with its own small or fine-tuned language model. This specialization would allow each agent to reason more efficiently within its domain and further reduce token usage while preserving collaborative reasoning.
\bibliographystyle{IEEEtran}
\bibliography{references_abbr}

\end{document}